# A new approach for evaluating internal cluster validation indices

Zoltán Botta-Dukát

Institute of Ecology and Botany, Centre for Ecological Research, Vácrátót

botta-dukat.zoltan@ecolres.hu

**Abstract**

A vast number of different methods are available for unsupervised classification. Since no algorithm and parameter setting performs best in all types of data, there is a need for cluster validation to select the actually best-performing algorithm. Several indices were proposed for this purpose without using any additional (external) information. These internal validation indices can be evaluated by applying them to classifications of datasets with a known cluster structure. Evaluation approaches differ in how they use the information on the ground-truth classification. This paper reviews these approaches, considering their advantages and disadvantages, and then suggests a new approach.

**Introduction**

Clustering is an unsupervised method for pattern recognition by grouping objects (Xu & Wunsch 2009). A vast number of different methods were developed for this purpose (e.g., Legendre & Legendre 1998; Podani 2000; Xu & Wunsch 2009). Different clustering approaches usually lead to different clusters of data. Different settings of parameters (including the number of clusters) generate further variation among clustering solutions. Since no algorithm performs best in all types of data (Pal & Biswas 1997), there is a need for a procedure for selecting the actually best-performing algorithm and parameter settings. This procedure is called cluster validation (Xu & Wunsch 2009).

Cluster validation may apply external or internal criteria to measure the "quality" of clustering. External criteria are calculated using additional information not used in the clustering. The most typical example is the similarity to the ground truth classification, but predictive power for any external variable is also applied. Internal criteria measure the "quality" of clustering based on variables used in clustering, focusing on compactness and separation of clusters or good representation of distances among objects (Handl et al. 2005).

Assuming that there is a single "true" grouping of a given dataset, clusterings are better or worse depending on how close they are to the "true" grouping (Akhanli & Hennig 2020). If this "true" grouping – often called ground truth – is known, external validation can be done. Typically, the ground truth is unknown, and internal validation has to be applied. Many different internal cluster validity indices (CVI) exist (for review, see, e.g., Milligan 1981; Vendramin et al. 2010). Thus, it is a challenging task for the user to choose one of them. Therefore, comparing the performance of internal CVI has been a recurring task since the 1980s (Milligan 1981; Milligan & Cooper 1985; Dimitriadou et al. 2002; Vendramin et al. 2010; Arbelaitz et al. 2013; Gagolewski et al. 2021).

Different attempts applied different approaches for evaluating the performance of CVIs. This paper aims to review and compare these approaches considering their pros and cons. Based on the lesson learned from this review, I suggest a new approach. The remainder of this paper is organized as follows. First, I shortly introduce the dataset, algorithm, and CVIs, which will be used in the illustrative examples. The next section is a review of former approaches, followed by the explanation of the proposed new approach and conclusions.

**Dataset, clustering algorithm, and CVIs used in the illustrative examples**

To illustrate the properties of different approaches, four internal validation indices were calculated for partitions of the Wine dataset downloaded from the UCI Machine Learning Repository (Dua & Graff 2017). This dataset consists of data from a chemical analysis of 178 wines grown in the same region in Italy but derived from three different cultivars. The chemical analysis determined the quantities of 13 constituents found in each of the three types of wines. All variables were standardized before analyses, and then wines were classified by the UPGMA algorithm using Euclidean distance. Partitions were created by cutting the dendrogram at different heights. Adjusted Rand-index (Hubert & Arabie 1985), the Calinski-Harabasz index (Calinski & Harabasz 1974), Davies and Bouldin index (Davies & Bouldin 1979), point biserial correlation (Milligan 1981), and mean silhouette width (Rousseeuw 1987) were calculated for each partition. All calculations were done in the R environment (R Core Team 2022), using *cluster* (Maechler et al. 2022), *mclust* (Scrucca et al. 2016), and *NbClust* (Charrad et al. 2014) packages.

**Critical evaluation of evaluation approaches**

Papers with comprehensive evaluations of cluster validation indices are listed in Table 1. Although almost all studies used a slightly different approach – the only exception is Dimitriadou et al. (2002), who followed the method of Milligan and Cooper (1985) - there are some common steps. All evaluation approaches use labeled data sets, where the ground truth (or reference) grouping is known. Labeling may come from the knowledge of the processes creating the dataset or from expert judgment. In the latter case, alternative reference groupings may exist (Gagolewski et al. 2021). These datasets are analyzed by different clustering methods to create a set of partitions (evaluation set). If hierarchical clustering is applied, the resulting tree has to be converted into a partition (or series of partitions). Hereafter, groups in the ground truth grouping will be called classes, while groups created by clustering algorithms will be called clusters. Accordingly, we will distinguish classification (a reference grouping) and clustering (a grouping that could be evaluated by CVI). After creating the set of partitions, the evaluated CVI is calculated for each partition. Approaches differ in the subsequent steps, where CVI values are contrasted with the reference classification. Often, but not in all approaches, choosing the partition with the highest (or lowest if a lower CVI value indicates a better partition) CVI is the next step. We will refer to this partition as the "optimal partition" and the number of clusters in this partition as the "optimal number of clusters".

Milligan & Cooper's almost forty-year-old paper (Milligan & Cooper 1985) is still the most highly cited among comprehensive evaluations (Table 1). It is surprising because later works (Vendramin et al. 2010; Gurrutxaga et al. 2011; Arbelaitz et al. 2013) strongly criticized their approach. Moreover, their approach is restricted to using CVIs to decide the optimal cluster number in hierarchical clustering. (Reflecting this aim, they use the term "stopping rule" in their paper.) They assumed that the optimal number of clusters should equal the number of classes (the right number of clusters). They ranked the compared CVIs based on the proportion of successes, counting each case as a 'failure' where the optimal number of clusters differ from the number of classes. The main criticism of this approach (Vendramin et al. 2010; Gurrutxaga et al. 2011; Arbelaitz et al. 2013) is that partition with the right number of clusters may considerably differ from the reference classification. On the other hand, a partition with the wrong number of clusters may be pretty similar to the reference classification. Note that the authors themselves have already recognized this possible bias. Therefore, beyond the evaluated internal CVIs, they calculated the similarity of partitions to the reference classification. In the great majority of the cases, the partition with the right number of groups was the most similar to the partition. Vendramin et al. (2010) pointed out that this good agreement between the right number of clusters and the highest similarity to the reference classification is instead an exception due to the well-separated classes than a general rule. In our example, the similarity to reference classification is maximal for eight clusters (adjusted Rand-index=0.793), while for three clusters, the adjusted Rand index is -0.011, worse than the random expectation (Figure 1).

Gurrutxaga et al. (2011) proposed a modification to eliminate this bias. They defined 'success' as the cases when the optimal partition is the most similar to the reference classification. This approach was applied by Arbelaitz et al. (2013). While the approach of Milligan and Cooper (1985) can be used only for comparisons within clustering algorithms, the approach of Gurrutxaga et al. (2011) allows for comparing clustering algorithms too.

The common feature of these two approaches is that the CVIs are characterized by the number of successes. Within failures, they do not distinguish the cases when the optimal partition is similar to and completely different from the reference classification.

In the classification of the Wine dataset, the optimal number of clusters is 8, 12, 8, and 2, according to the Calinski-Harabasz index, point biserial correlation, mean silhouette width, and Davies-Bouldin index, respectively (Figure 2, Table 2). Thus, both Davies and Bouldin index and point biserial correlation would fail according to either of the two approaches. However, they proposed completely different solutions as optimal: the solution with twelve clusters is rather similar to the ground truth classification (adjusted Rand-index is 0.775), while the partition with only two clusters is completely different (adjusted Rand-index is -0.002) (Figure 1).

Milligan and Cooper (1985) gave a breakdown of failures according to the difference between the number of clusters in the optimal partition and the number of classes. It assumes that partitions of a dataset with four classes into three and five clusters are equally wrong. Vendramin et al. (2010; Figs 1-3) illustrated that this assumption does not hold.

In his pioneering work, Milligan (1981) proposed that ranking partitions by a good internal CVI should be correlated with their ranking based on similarity to ground truth classification. Internal CVIs have a maximum (or minimum) as a function of the number of clusters. Milligan presumed that external indices might show a monotonic trend, which hinders the high (rank) correlation between external and internal indices when the number of clusters varies among partitions. Therefore, he restricted the evaluation set to partitions with the same number of clusters. Probably, this conjecture is why another approach was applied four years later by Milligan and Cooper (1985). Interestingly, Milligan and Cooper (1986) themselves showed later that although unadjusted Rand, Jaccard, or Folwkes-Mallows indices showed a monotonic pattern for data without cluster structure, adjusted Rand index had no trend in this case and had a clear peak for datasets with a cluster structure. Based on these results, Vendramin et al. (2010) argued that rank correlation can be calculated over partitions with different numbers of clusters. They stated that this approach is advantageous because it uses more information, not only the optimal partition but also the value of CVI in all partitions. However, this advantage can be questioned. End-users of CVIs typically use them to choose the best partition and do not worry about how other partitions are ranked. In the analysis of the Wine dataset where both the Calinski-Harabasz index and average silhouette width choose the same optimal partition (i.e., partition most similar to the ground truth), but they considerably differ in correlation with the adjusted Rand index (Table 2). Thus, according to the approach of Vendramin et al., the Calinski-Harabasz index is much better. Its reason is that partition with two clusters is the second best solution according to the mean silhouette width, while the Calinski-Harabasz index correctly recognizes that the partition with two clusters is wrong. Moreover, according to the approach of Vendramin et al., point biserial correlation is the best of the four indices (Table 2), although it chooses partitions with 12 clusters as optimal. Note that correlation with the adjusted Rand index depends on the set of partitions used in the evaluation (Figure 3).

A common feature of the above-discussed indices is that they assume CVI will be used for choosing the best one from the given set of partitions. Gagolewski et al. (2021) regard CVI as a function that should be optimized to find the optimal partition. Therefore, in their approach, not partitions created irrespectively to CVI are ranked according to CVI value, but partitions are created by optimizing a CVI. In this case, the higher the similarity between the final partition and the ground truth, the better the CVI. Since CVI values are optimized, the maximal goodness is always one. When CVI values are applied to evaluate classifications, the maximum similarity to the ground truth may considerably vary among datasets and depending on the considered classification methods.

**A new approach for evaluating internal cluster validation indices**

Here I propose a new approach for evaluating CVIs that eliminates the drawbacks mentioned above. My starting point is that the only information the end-users will use is which partition was selected by CVI as the best. If this partition is the most similar partition to the ground truth, the CVI is maximally good. If not, the goodness should depend on the similarity of the selected partition to the ground truth and its maximum in the set of partitions actually considered.

The new approach consists of the following steps:

1. Create several partitions for a dataset with ground truth classification. The partitions may differ in the algorithm used to create them and their number of clusters.
2. Calculate the similarity of each partition to the ground truth. Denote $S(i)$ the similarity of partition $i$.
3. Calculate CVI for each partition. Denote the value of the index for partition $i$ by V(i).
4. Denote the best partition by $b$. $b$=argmax{$V$(i)} or $b$=argmin{$V$(i)} depending on if the maximum or minimum of the index indicates the best partition.
5. Goodness of the cluster validation index is $S(b)/\max\{S(i)\}$

The possible maximum of this goodness is one, irrespective of the set of partitions used in the evaluation. Therefore, its value can be compared between datasets. Thus, calculating the mean or the median over several datasets characterizes the performance of the cluster validation index by a single number. Selecting the best subset of cluster validation indices based on non-parametric tests (Arbelaitz et al. 2013) is preferable instead of choosing the single best CVI. An example of such an approach is Pakgohar et al. (in prep).

Applying this new approach to the example of the Wine dataset (Table 2) shows that the Calinski-Harabasz index and average silhouette width selected the best partition. Point biserial correlation also performed well, while the Davies-Bouldin index reached small goodness.

This ranking of cluster validation indices is considerably different from rankings based on both approaches of Milligan & Cooper (where all indexes failed) and Vendramin et al. (where point biserial correlation is the best, and average silhouette width performed wrong) (Table 2). Evaluation using the approach of Gurrutxaga et al. gave the most similar results. However, it did not distinguish the performance of the point biserial correlation and the Davies-Bouldin index.

**Conclusions**

The new approach proposed in this paper has three advantageous properties: (1) it measures goodness in a continuous instead of binary (success/fail) scale; (2) it focuses on the quality of partition selected by the index; (3) its value is not sensitive to the range of cluster numbers. Note that the last property only holds if changing the range does not influence the identity of the best partition. I consider the second property is advantageous because end-users often focus on the best partition and neglect the value of the index in other partitions. When this condition does not hold, the approach of Vendramin et al. should be applied.

**Table 1**

|  |  |
|---|---|
| (Milligan & Cooper 1985) | 5002 |
| (Arbelaitz et al. 2013) | 1177 |
| (Milligan 1981) | 554 |
| (Dimitriadou et al. 2002) | 399 |
| (Vendramin et al. 2010) | 388 |
| (Gurrutxaga et al. 2011) | 82 |
| (Gagolewski et al. 2021) | 10 |

**Table 2.** Evaluation of the four CVIs based on the analysis of example dataset using different approaches

|  | Best value of the index | Number of clusters at the best value of the index | adjusted Rand index for the best partition | Goodness according to | | | | |
|---|---|---|---|---|---|---|---|---|
|  |  |  |  | Milligan and Cooper | Gurrutxaga et al. | Vendramin et al. considering partitions with cluster number | | the new approach |
|  |  |  |  |  |  | from 2 to 15 | from 2 to 100 |  |
| Calinski-Harabasz | 29.9257 | 8 | 0.7929 | 0 (fail) | 1 (success) | 0.9031 | 0.7145 | 1 |
| point biserial correlation | 0.6832 | 12 | 0.7755 | 0 (fail) | 0 (fail) | 0.9517 | 0.9739 | 0.9780 |
| avg. silhouette width | 0.2719 | 8 | 0.7929 | 0 (fail) | 1 (success) | 0.4990 | -0.6686 | 1 |
| Davies-Bouldin | 0.6091 | 2 | -0.0020 | 0 (fail) | 0 (fail) | -0.1015 | 0.5754 | -0.00245 |

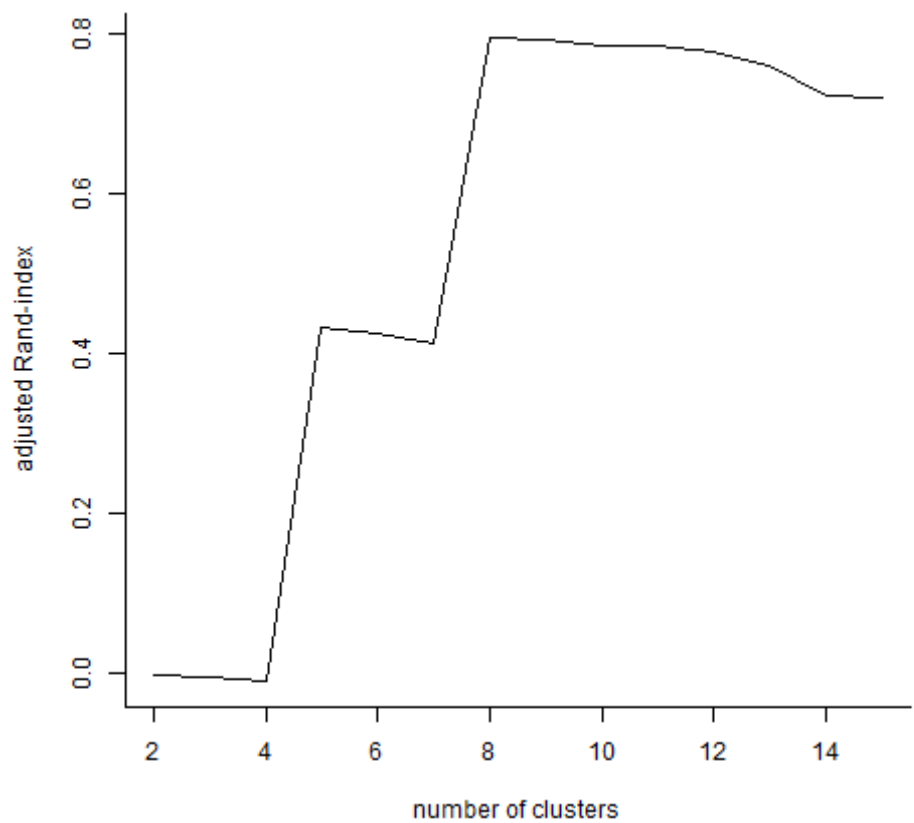

Figure 1. The similarity between reference classification and partitions with different clusters resulted from cutting the dendrogram at different heights. The similarity is highest at eight clusters.

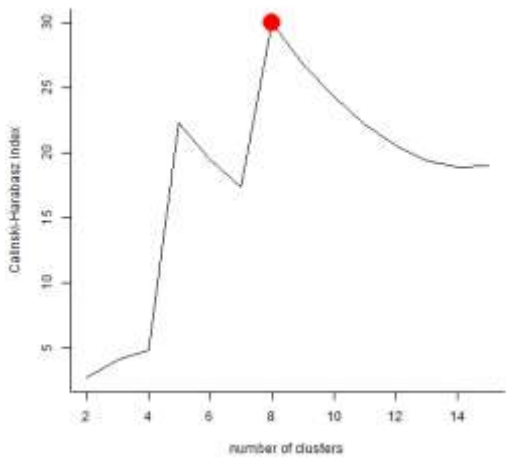
a)

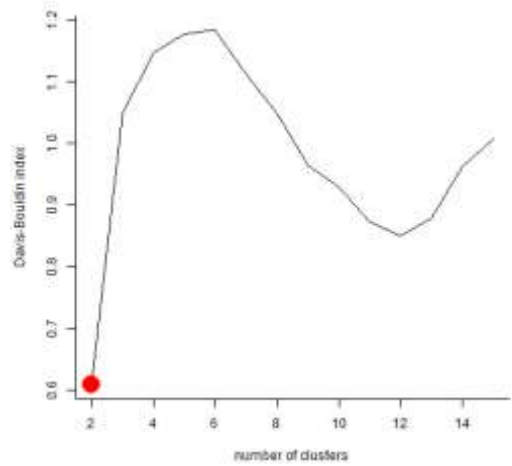
b)

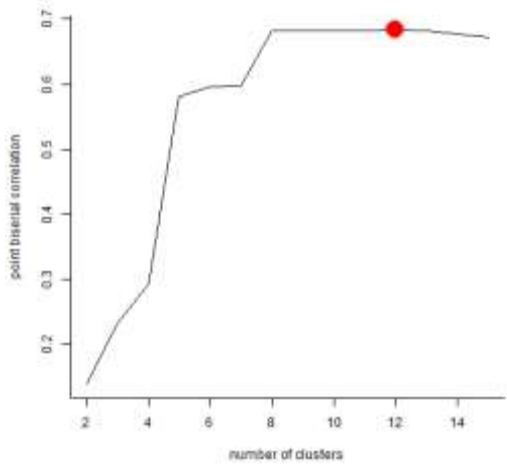
c)

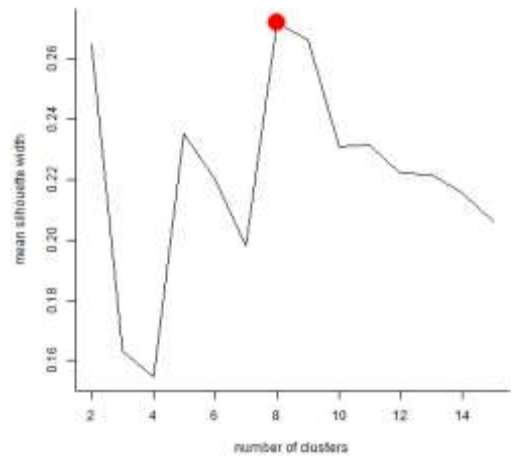
d)

Figure 2. Values of four internal CVIs for partitions with different numbers of clusters resulted by cutting the dendrogram at different heights. Red dots indicate the best grouping (lowest value for the Davis-Bouldin index and highest value for the other three indices).

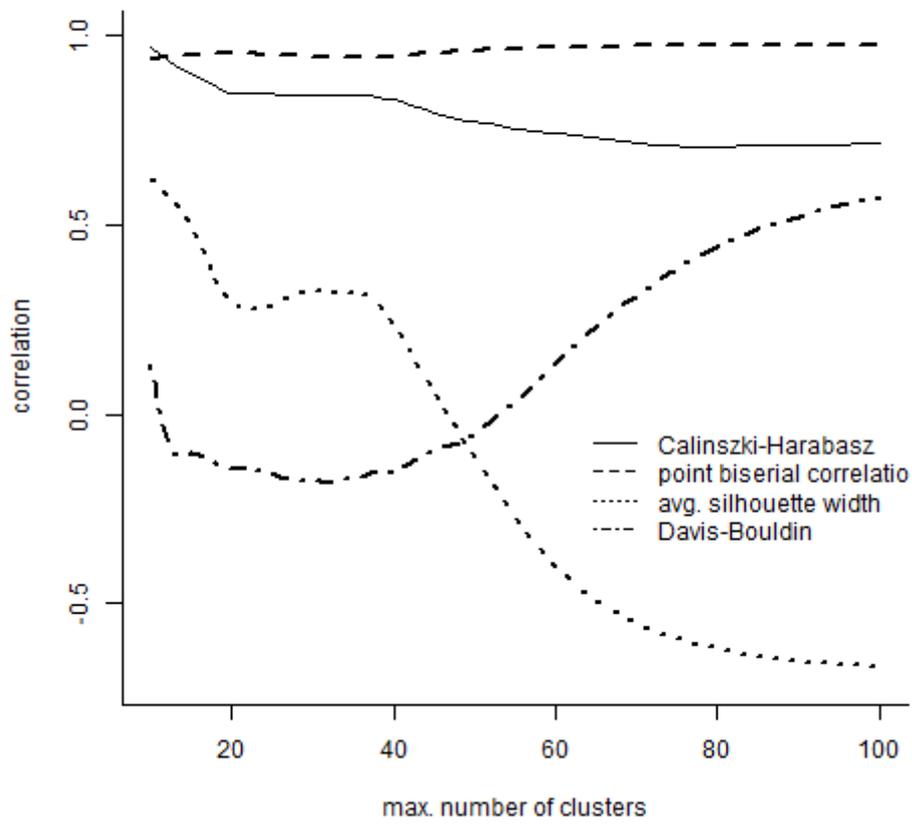

Figure 3. Correlation between internal CVIs and similarity to reference classification (measured by adjusted Rand index) depends on the set of considered partitions. In this example, the lowest number of clusters was always two, while its maximum changed from 10 to 100.